\documentclass{article}

\PassOptionsToPackage{numbers, compress}{natbib}

\usepackage[preprint]{neurips_data_2024}

\usepackage{tikz}
\usepackage{adjustbox}
\usepackage{xcolor}
\usetikzlibrary{positioning}

\usepackage[title]{appendix}

\usepackage{wrapfig}

\usepackage{hyperref}       
\usepackage{url}            
\usepackage{booktabs}       
\usepackage{amsfonts}       
\usepackage{nicefrac}       
\usepackage{xcolor}         
\usepackage{enumitem}

\usepackage{longtable}
\usepackage{booktabs}
\usepackage{array}
\usepackage{pbox}

\usepackage{listings}
\usepackage{setspace}

\usepackage{tikz}
\usepackage[acronym]{glossaries}

\lstdefinestyle{plantuml}{
  language      = {},
  basicstyle    = \small\ttfamily,
  keywordstyle  = \color{teal!70!black}\bfseries,
  commentstyle  = \color{gray!70},
  stringstyle   = \color{orange!80!black},
  showstringspaces = false,
  columns       = fullflexible,
  keepspaces    = true,
  breaklines    = true,
  frame         = single,
  framerule     = 0.4pt,
  rulecolor     = \color{black},
  backgroundcolor = \color{gray!5},
}
\usetikzlibrary{positioning, arrows.meta, shapes.multipart, calc}

\title{On the Evaluation of Engineering\\ Artificial General  Intelligence}

\newacronym{cae}{CAE}{Computer Aided Design}

\newcommand{\name}{\texttt{eAGI}}

\newcommand{\comment}[1]{\ignorespaces}

\author{%
Sandeep Neema \quad 
Susmit Jha \quad 
 Adam Nagel \quad 
  Ethan Lew \And
   Chandrasekar Sureshkumar \quad 
Aleksa Gordic \quad
 Chase Shimmin \quad
 Hieu Nguygen \And
 Paul Eremenko \\
 \\
  P-1 AI, Inc., San Francisco, CA, Email: \texttt{sandeep@p-1.ai} \\
}

\begin{document}

\maketitle

\begin{abstract}
We discuss the challenges and propose a framework for evaluating engineering artificial general intelligence (\name) agents.
We consider \name\ as a specialization of artificial general intelligence (AGI), deemed capable of addressing a broad range of problems in the engineering of physical systems and associated controllers. We exclude software engineering for a tractable scoping of \name\ and expect dedicated software engineering AI agents to address the software implementation challenges. 
Similar to human engineers, \name\ agents should possess a unique blend of background knowledge (recall and retrieve) of facts and methods, demonstrate familiarity with tools and processes, 
exhibit deep understanding of industrial components and well-known design families, and be able to engage in creative problem solving (analyze and synthesize), transferring ideas acquired in one context to another . 
Given this broad mandate, evaluating and qualifying the performance of \name\ agents is a challenge in itself and, arguably, a critical enabler to developing \name\ agents.
In this paper, we address this challenge by proposing an extensible evaluation framework that specializes and grounds Bloom's taxonomy - a framework for evaluating human learning that has also been recently used for evaluating LLMs - in an engineering design context.
Our proposed framework advances the state of the art in benchmarking and evaluation of AI agents in terms of the following: (a) developing a rich taxonomy of evaluation questions spanning from methodological knowledge to real-world design problems;
(b) motivating a pluggable evaluation framework that can evaluate not only textual responses but also evaluate structured design artifacts such as CAD models and SysML models; and 
(c) outlining an automatable procedure to customize the evaluation benchmark to different engineering contexts. 
\end{abstract}

\section{Introduction}

Engineering the physical world remains among the most cognitively and technically demanding human endeavors. Designing aircraft, electric vehicles, energy systems, robots, or climate control infrastructure requires not only a deep understanding of physical laws but also the ability to model, analyze, and optimize systems that are tightly coupled across disciplines. Unlike software engineering, where modularity and abstraction have enabled powerful automation, physical systems engineering remains largely manual, iterative, and expertise-driven. Engineers must reason across thermal, structural, electrical, and control domains; often with incomplete data, uncertain constraints, and evolving design goals.
To mitigate this complexity, the industry has embraced tools such as 
Computer-Aided Engineering (CAE) and Model-Based Systems Engineering (MBSE) frameworks. These tools simulate physical behavior, guide architecture decisions, and improve verification workflows. Yet their adoption is limited by steep learning curves and reliance on domain-specific heuristics. Even with these tools, design iteration remains time-consuming and error-prone, especially when novel architectures or trade-offs must be explored.

Recent advances in AI and deep learning, particularly in Large Language Models (LLMs) and agentic AI frameworks, offer promising avenues to reduce this burden. Early applications of LLMs include code generation for simulation scripts, language interfaces for CAE tools, and documentation assistants. However, these tools essentially serve as procedural copilots. They improve interaction efficiency but do not exhibit the kind of domain reasoning, multi-physics modeling, or design iteration that is necessary for true engineering collaboration.

To realize transformative gains, role of AI in engineering must evolve beyond interface-level assistance and become an active, reasoning partner in the design process. This evolution requires a new class of AI agents -- \name\ agents -- that can engage in creative problem-solving, simulate physical systems, evaluate trade-offs, and produce artifacts such as structured designs, models, or control strategies. \name\ agents are not just retrieval systems; they must be capable of reasoning with constraints, synthesizing cross-domain knowledge, and adapting learned principles to novel problems.

A central challenge in developing such systems is the absence of robust, structured, and context-aware methods for evaluating engineering intelligence. Unlike general NLP benchmarks or vision tasks with fixed outputs, engineering problems are context dependent and often have no single ``correct" answer. Performance must be assessed across different levels of reasoning and creativity, taking into account how well an \name\ agent understands the design context, satisfies system constraints, and explores the solution space across multiple domains of physics and engineering.

\section{Engineering Artificial General Intelligence (\name)}
{\name} aims at 
cognitive automation of what a human engineer does, that is,  
 {\name} is a specialization of AGI  designed to perform expert-level reasoning, analysis, and creative synthesis for physical systems engineering. Similar to human engineers, \name\ agents must combine factual and methodological knowledge with tool fluency, domain-specific modeling intuition, and adaptive, iterative, reasoning-guided design thinking. 
An \name\ agent should exhibit the following characteristics:
\begin{itemize}[leftmargin=*]
    \item {Background knowledge and recall:} It should retrieve relevant engineering laws, formulas, heuristics, and design rules when prompted.
    \item {Process and tool familiarity:} It should understand workflows in CAE and MBSE environments and be able to interface with simulation pipelines or design models. It should be aware of available tools, understand their limitations, configuration requirements, and constraints, be able to generate valid inputs to these tools, and ingest their output. 
    \item {Contextual understanding:} It should recognize components, systems, and recurring patterns from industry-standard designs in a domain, such as heat exchangers or propeller assemblies.
    \item {Creative and adaptive reasoning:} It should explore novel design solutions, adapt knowledge to unfamiliar systems, and optimize performance across competing objectives.
    \item {Collaborate and communicate:} It should participate and collaborate with human engineers in design activities through communication that is natural, domain, and context-aware. 
\end{itemize}

The goal of eAGI is cognitive automation, so we explicitly exclude surrogate modeling and approximating detailed simulation or analysis tools and instead expect eAGI to be able to call such tools or their machine learning accelerated surrogates. We expect eAGI to be able to distill key design drivers from requirements, postulate one or more solutions, and conduct first-order design sizing using tools when needed. 
This multifaceted skillset differentiates \name\ from both narrow AI applications in engineering (e.g., automated meshing, parametric sweeps) and general-purpose LLMs (use in generative and predictive auto-completion of artifacts). While current LLMs excel at natural language understanding and code generation, they fall short in areas where structured physical modeling, precise constraints, qualitative and quantitative cross-domain causal reasoning is required.

Given the breadth of tasks expected of an \name\ agent, \textit{evaluating its capabilities becomes a foundational challenge}. Existing benchmarks either test shallow factual knowledge (e.g., textbook questions) or rely on rigid, domain-specific problems with fixed outputs. Neither of these approaches captures the range of reasoning, modeling, and synthesis skills needed for effective engineering intelligence.
We propose a comprehensive evaluation framework for \name\ agents that:
\begin{enumerate}
[leftmargin=*]
    \item Specializes Bloom’s taxonomy of human learning outcomes for engineering contexts, providing a structured hierarchy of cognitive abilities from factual recall to reflective, cross-domain design.
    \item Introduces a {rich taxonomy of evaluation questions}—spanning from low-level methods to high-level system synthesis requiring end-to-end problem solving, that can be instantiated across multiple engineering problem domains.
    \item Supports {pluggable and adaptive benchmark customization}, where evaluations can be dynamically generated and tailored to different systems (e.g., eVTOL propulsion, HVAC design) and physics domains (e.g., thermal, electrical, structural).
\end{enumerate}

Our framework is designed to evolve alongside the growing capabilities of \name\, offering a scalable and extensible methodology to quantify engineering competence. It lays the groundwork for systematic development through principled qualification and objective benchmarking of AI agents that aim to collaborate meaningfully in the engineering of the physical world.

\label{sec:intro}

\section{Related Work}
\textbf{General LLM and Agent Benchmarks.} Evaluating Large Language Models (LLMs) has traditionally focused on general benchmarks and natural language tasks such as language comprehension, reasoning, and code generation. These standard evaluations assess performance on tasks like multiple-choice question answering (MMLU\cite{hendrycks2020mmlu}), reading comprehension (SuperGLUE\cite{wang2019superglue}), and logical reasoning (BIG-Bench Hard\cite{srivastava2022beyond}). The focus of these evaluations is on broad linguistic and logical capabilities across multiple domains. The most commonly used benchmarks include: MMLU (Massive Multitask Language Understanding), which assesses factual knowledge and reasoning across 57 domains \cite{hendrycks2020mmlu}; SuperGLUE, which evaluates natural language understanding via reading comprehension, textual entailment, and coreference resolution \cite{wang2019superglue}; and BIG-Bench Hard (BBH), which measures advanced reasoning, common sense, and problem-solving abilities \cite{srivastava2022beyond}. Other evaluations such as GSM8K\cite{cobbe2021gsm8k} and MATH\cite{hendrycks2021math} test mathematical problem-solving ability . 

While these benchmarks offer insights into general intelligence, they fail to capture the depth of knowledge, reasoning, and application required in specialized fields like engineering, medicine, and law. The LLM agents used in niche domains such as engineering require more than just linguistic proficiency, logical reasoning, and general problem-solving abilities, and new evaluation methodologies must be designed to measure their domain-specific capabilities accurately.

\textbf{Domain-Specific Benchmarks.} 
Recent efforts have focused on constructing benchmarks that assess LLMs and autonomous agents in specialized domains. In software engineering, SWE-bench \cite{swebench} offers a challenging benchmark based on real-world GitHub issues and pull requests. It evaluates whether models can propose code edits that resolve actual bugs or feature requests in open-source repositories. The benchmark demands not just code generation but also comprehension of issue descriptions, relevant file locations, and dependency management. Similarly, Devin \cite{devin}, introduced as an autonomous AI software engineer, is evaluated on tasks including full-stack debugging, implementation of GitHub issues, and multi-step planning for end-to-end software development. Complementing these large-scale, real-world tasks are programming-focused benchmarks such as LeetCode~\cite{coignion2024performance} and Codeforces~\cite{li2022competition}, which are widely used to evaluate algorithmic problem-solving capabilities. LeetCode-style problems assess models on standard data structures, algorithms, and computational thinking, while Codeforces offers dynamic, time-constrained competitive programming challenges that test both reasoning under pressure and syntactic correctness. These coding platforms provide task variety and difficulty scaling, making them suitable for fine-grained analysis of code synthesis and logic generalization. In medicine, benchmarks like MedQA \cite{jin2019medqa}, MedMCQA \cite{pal2022medmcqa}, and PubMedQA \cite{jin2019pubmedqa} evaluate LLMs on tasks ranging from clinical reasoning to biomedical literature comprehension. SciKnowEval \cite{zhang2024sciknoweval} assesses LLMs across STEM fields using a five-tier taxonomy: memorization, comprehension, reasoning, discernment, and application. In law, evaluations such as LexGLUE \cite{chalkidis2022lexglue} and LawBench \cite{fei2023lawbench} focus on statutory interpretation, legal document comprehension, and judicial reasoning. 
Benchmarks have also been proposed to evaluate LLMs' understanding of engineering documentation~\cite{doris2025designqa}, and existing LLMs have been found to 
have several key deficiencies, such as the inability to provide reliable output, a lack of relevant domain knowledge, unawareness of industry standards, and failure to integrate seamlessly with existing workflow~\cite{alam2024automation}. 
These domain-specific benchmarks move beyond generic linguistic metrics and assess expert-level competencies in realistic scenarios. MMLU-Pro \cite{wang2024mmlu}, an extension of MMLU, expands into applied engineering and physics problems, addressing some of the limitations of traditional benchmarks for evaluating technical expertise. 

\textbf{AGI and General Intelligence Evaluation.} 
Evaluating Artificial General Intelligence (AGI) requires metrics that transcend domain-specific skillsets and capture core attributes of generalization, transfer, memory, planning, and agency. Proposed evaluation frameworks for AGI often rely on multi-task performance, real-world decision-making simulations, and generalization under uncertainty \cite{goertzel2014artificial}. Cognitive architectures have been tested in lifelong learning scenarios~\cite{romero2023perspective}, where models must adapt and improve over time without forgetting past tasks. Others have proposed meta-evaluation setups, such as ARC-Challenge~\cite{chollet2024arc} tasks, that probe whether a system can reason or decompose long-horizon goals. Beyond standard task performance, AGI evaluation must also consider agency (planning, tool use), robustness (adversarial perturbation), and autonomy (long-term task completion with minimal intervention). 
Morris et al.~\cite{morris2023levels} propose a framework for evaluating progress toward Artificial General Intelligence (AGI) based on two key axes: performance (depth of capabilities) and generality (breadth across tasks). Analogous to levels of autonomous driving, their ``Levels of AGI" taxonomy introduces a matrix that categorizes AI systems from Emerging to Competent to Expert to Virtuoso to Superhuman. The authors argue for operational definitions of AGI that prioritize capabilities over underlying mechanisms. They further distinguish between capability and autonomy, proposing a separate set of autonomy levels ranging from tool, consultant, collaborator, expert, and agent. This framework provides a structured approach to evaluating and benchmarking AGI systems along a continuum, rather than using binary thresholds. 
Despite advances, no widely accepted standard currently exists for AGI evaluation, and many existing benchmarks fail to distinguish between memorized responses and genuinely novel reasoning. By focusing on the engineering domain where artifacts can be objectively evaluated, we present a principled approach to define different levels of \name\ and propose an \name\ evaluation framework.

\textbf{Skill Progression Models.} Several cognitive development models have been proposed to describe how humans acquire domain expertise over time, particularly in complex professions like engineering, medicine, and law. Among these, the Dreyfus Model of Skill Acquisition \cite{dreyfus1986mind} is one of the most widely cited, outlining a five-stage continuum from Novice to Expert, where practitioners shift from rule-based responses to context-sensitive, intuitive action. Similarly, models such as the Benner Model~\cite{benner1984novice} (adapted from Dreyfus for clinical nursing), the Rasmussen Skill-Rule-Knowledge (SRK) framework~\cite{rasmussen1983skills}, and the Berliner Model~\cite{berliner2004expert} in pedagogy all emphasize the transition from declarative knowledge to procedural and tacit knowledge, relying heavily on experiential learning and situated practice. These models are effective for evaluating human professionals, but present limitations when applied to LLMs or agentic systems: they assume continuous exposure and feedback loops.
In contrast, Bloom’s Taxonomy \cite{bloom1956taxonomy} provides a hierarchical, task-oriented structure that is far more amenable to evaluating AI systems. 
Unlike the Dreyfus model and its variants, which require subjective evaluation of behavior over time, Bloom’s framework lends itself to quantitative, benchmarkable assessments. 
Therefore, for the purpose of evaluating \name\, Bloom’s Taxonomy offers a more practical, granular, and scalable foundation, enabling the construction of benchmarks that both reflect domain-relevant challenges and allow for comparisons across models and systems.

\textbf{Bloom’s Taxonomy for Hierarchical Evaluation.} 
A promising approach to evaluating LLMs, particularly in specialized or safety-critical domains, draws from educational psychology frameworks such as Bloom’s Taxonomy. Originally designed to classify educational goals, Bloom’s Taxonomy has been adapted to measure LLM capabilities across cognitive levels, including knowledge recall, comprehension, application, analysis, synthesis, and evaluation \cite{bloom1956taxonomy}. Recent benchmarks, such as SciKnowEval \cite{zhang2024sciknoweval} and LawBench \cite{fei2023lawbench} have implicitly or explicitly incorporated hierarchical evaluation frameworks. For instance, SciKnowEval assesses memory, reasoning, and application skills in science domains; LawBench categorizes legal tasks into memorization, comprehension, and application. This hierarchical perspective enables more granular insights into model competence and aligns evaluation with human cognitive development. Adapting Bloom's framework to engineering intelligence opens avenues for more structured assessments - ranging from accurate recall of facts and methods to the design and optimization of complex systems under constraints.

\label{sec:rel}

\section{Engineering Cognition Levels Using Bloom's Taxonomy}
Engineering design (Figure~\ref{fig:eng}) is the process of translating requirements and constraints into realizable physical systems. This process involves selecting and configuring components and defining their topology, that is, how components are connected logically and arranged in space and function within the context of a system-level objective.
Engineering problems are often underspecified and open ended such that they do not have a single correct solution. Instead, they span a rich space of possible configurations, each governed by various qualitative and quantitative constraints. 
These constraints can range from fundamental physical laws (like energy conservation or structural mechanics) and performance targets (such as thrust, efficiency, or reliability) to discrete choices (like selecting components and preferred reuse of existing subsystems), spatial arrangements (e.g., heat exchanger layouts, wiring harness routing), manufacturing limitations (e.g., tolerances),  and adherence to design standards and regulations (including safety). The core challenge lies in creating a design that meets all requirements and performance objectives. This design must then be assessed for its feasibility, robustness, alignment with human and organizational preferences, and optimality under both real-world and simulated scenarios.

We identify three major dimensions to measure the complexity of engineering problems.
\begin{enumerate}
    \item \textbf{Directionality:} The first dimension relates to whether the problem requires a forward analysis to evaluate performance and conformance to requirements, or whether it needs inverse reasoning and synthesis from requirements to design. Since multiple designs are feasible for the same set of requirements and most engineering evaluation tools only provide the forward evaluation of a design, the inverse direction of the synthesis is typically ill-posed and more challenging.
    \item \textbf{Design Behavior:} The second dimension is the level of details of the behavior of the engineering design. Typical designs are inherently multiphysics, and design evaluation involves running dedicated tools for specialized tasks such as electrical analysis, thermal analysis, computational fluid dynamics, etc. Whether this analysis is focused on the static aspect of the design, such as structure, or whether it focuses on its dynamic behavior significantly changes the complexity of the engineering tasks. Typically, dynamic behavioral analysis is more challenging than static analysis of designs.  
    \item \textbf{Design Scope:} The third dimension is whether the synthesis or evaluation problem is open world or closed world. In the simpler closed-world setting, the space of design decisions is bounded; for example, the design could be partially specified as a template with some remaining details to be filled out or an existing design needs to be updated to meet a regulatory standard. Similarly, for evaluation and analysis tasks, the open-world setting would require the engineer to identify all the relevant performance metrics for a problem. For example, the vibration analysis of a design might depend on material choice, and an experienced engineer can identify what analysis is needed for a design. The open-world design and analysis problems are naturally more difficult than the bounded design and evaluation. 
\end{enumerate} 
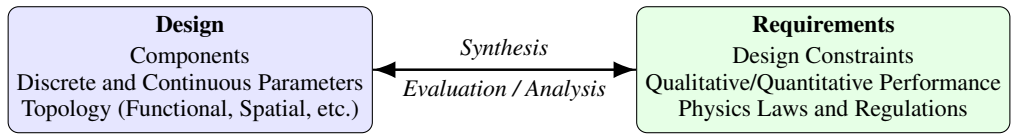
\begin{figure}[!ht]
\centering
\begin{tikzpicture}[
  node distance=1.8cm and 3.5cm,
  every node/.style={font=\small},
  box/.style={draw, rectangle, rounded corners, minimum width=4.8cm, minimum height=1.2cm, align=center},
  arrow/.style={-{Latex[length=3mm]}, thick},
  reverse/.style={-{Latex[length=3mm]}, thick},
  label/.style={font=\footnotesize\itshape}
]
\node[box, fill=blue!10] (design) {\textbf{Design}\\[2pt]
 \begin{tabular}{@{}c@{}}
    Components  \\
    Discrete and Continuous Parameters \\
    Topology (Functional, Spatial, etc.)
  \end{tabular}
};

\node[box, right=of design, fill=green!10] (requirements) {\textbf{Requirements}\\[2pt]
   \begin{tabular}{@{}c@{}}
    Design Constraints \\
    Qualitative/Quantitative Performance \\
    Physics Laws and Regulations
  \end{tabular}
};

\draw[arrow] (requirements) -- node[above, label] {Synthesis} (design);
\draw[reverse] (design) -- node[below, label] {Evaluation / Analysis} (requirements);

\end{tikzpicture}
\caption{The two key categories of tasks in engineering involve the {\it synthesis} of design from given requirements and the {\it evaluation or analysis} of a given design to measure its performance and compliance with constraints and regulations.}
\label{fig:eng}
\end{figure}
Engineering involves generating a design by selecting appropriate components, setting discrete and continuous parameters to optimal values, and coming up with the right functional and spatial topology to connect these components in an appropriate physical layout. This design is expected to satisfy requirements that could specify qualitative and quantitative performance objectives and constraints over design choices. The design is expected to be consistent with physics laws and compliant with regulatory requirements. Figure~\ref{fig:eng} illustrates the two key tasks in the field of engineering - synthesis and analysis. 

Factoring in these engineering problem complexity dimensions, we propose a 6-level hierarchy (Table~\ref{tab:eagi_complexity}) grounded in Bloom's taxonomy to characterize engineering cognition levels and systematically evaluate \name\ agents. 
These levels reflect ascending competencies—starting from factual recall and culminating in design synthesis, abstraction, and self-awareness. 
We use the same labels as the original Bloom's taxonomy where the highest level is evaluation and judgment, and the next level below is create or synthesis, that is the ability of creatively putting things together.  
We associate distinct engineering task types with each level as described below.

\begin{table}[h]
\centering
\small
\begin{tabular}{@{}lllll@{}}
\toprule
\textbf{\name\ Level} & \textbf{Directionality} & \textbf{Design Behavior} & \textbf{Design Scope} \\
\midrule
Level 1: Remember & Forward (Evaluation) & N.A. & Closed World  \\
Level 2: Understand & Forward & Static  & Closed-world \\
Level 3: Apply & Forward & Static + Dynamic & Closed-world \\
Level 4: Analyze & Forward + Partial Inverse & Static + Dynamic & Closed-world  \\
Level 5: Create & Forward + Inverse & Static + Dynamic & Semi-open-world  \\
Level 6: Reflect & Bidirectional & Static + Dynamic  & Fully open-world  \\
\bottomrule
\end{tabular}
\caption{Mapping of \name\ levels to three dimensions of engineering complexity: directionality (forward vs. inverse reasoning), behavior complexity (static or dynamic), and problem scope (closed vs. open). Each level requires increasing autonomy, generalization, and abstraction.}
\label{tab:eagi_complexity}
\end{table}

\vspace{-0.4cm}

\noindent \textbf{Level 1 (Remember) - Accurate Recall of Factual Information.} At this basic level, an \name\ agent must demonstrate reliable access to factual, domain-specific knowledge. This includes recalling definitions, equations, and engineering principles, accessing properties of standard components from model libraries, and citing regulatory standards or design codes.  Tasks at this level resemble look-up or Q\&A behavior but form the basis for more complex cognition. Accuracy and consistency are key metrics for evaluating the responses here.

\noindent \textbf{Level 2 (Understand) - Semantic Understanding of Designs.} This level assesses the AI’s ability to interpret and explain the structure and function of a given design. Core capabilities include identifying components in a schematic or model, understanding the spatial and functional topology of components within a system, and demonstrating an understanding of causal behavior between a design choice and observed performance. This level ensures that the model does not just recognize parts but understands their roles, interactions, and design intent.

\noindent \textbf{Level 3 (Apply) – Design Evaluation and Operational Reasoning.} An \name\ agent  at this level is expected to apply learned engineering principles to assess or manipulate designs. Capabilities include predicting the performance of new or unseen designs, making component substitutions while maintaining system feasibility, and invoking external tools such as solvers, simulators, or parametric models to evaluate or update design properties.
These tasks require the integration of reasoning, simulation outputs, and procedural operations - a significant leap from static understanding.

\noindent \textbf{Level 4 (Analyze) – Design In-filling and Error Diagnosis.} At this level, the system goes beyond evaluation to autonomously diagnose and resolve design issues. Typical tasks include completing partially specified designs using appropriate domain rules and constraints; detecting inconsistencies, errors, or violations in a given design; and proposing viable corrections or workarounds. This level embodies inverse reasoning from observed failures or incomplete designs back to underlying causes, mirroring human debugging in engineering workflows.

\noindent \textbf{Level 5 (Create) – Design Synthesis from Requirements.} This level involves the creative synthesis of new designs, especially under novel or shifted constraints. It includes generating full system designs from a given specification, adapting existing designs to new performance goals or regulatory requirements, and exploring the design space to approach or improve upon the Pareto frontier for trade-offs like performance vs. cost or weight vs. strength.
This level requires balancing competing objectives and leveraging domain knowledge in a generative form beyond pattern completion or memorized design templates.


\noindent \textbf{Level 6 (Reflection and Abstraction) – Meta-Reasoning and Critique.} At the highest level, an \name\ agent exhibits reflective and meta-cognitive abilities, including 
critiquing design decisions, 
inferring systemic implications, 
deriving and explaining general principles from multiple specific cases, and identifying situations where existing tools may be insufficient. \name\ can recognize its own uncertainty, limitations, or biases in design evaluation or synthesis.
This level reflects expert-level design reasoning, including judgment, abstraction, and the ability to explain and defend decisions - hallmarks of expert human engineers.

\begin{wrapfigure}{r}{0.36\textwidth} 
  \centering
\begin{tikzpicture}[
    every node/.style={
        rectangle,
        draw,
        rounded corners,
        font=\sffamily\small,
        align=center,
        minimum height=0.8cm
    }
]

\node[minimum width=3.5cm] (remember)  at (0,0)   {Remember};
\node[minimum width=3cm] (understand) at (0,0.8) {Understand};
\node[minimum width=2.5cm] (apply)      at (0,1.6)   {Apply};
\node[minimum width=2cm] (analyze)    at (0,2.4) {Analyze};
\node[minimum width=1.5cm] (create)     at (0,3.2)   {Create};
\node[minimum width=1cm] (reflect)    at (0,4) {Reflect};
\end{tikzpicture}
   \caption{6 \name\ Levels based on Bloom's taxonomy} 
\end{wrapfigure}
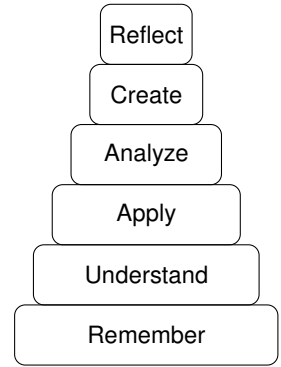
These levels can be located along the three dimensions for measuring the complexity of engineering problems identified earlier. With respect to forward (analysis) vs reverse (synthesis), Levels 1-3 operate predominantly in the forward direction of evaluating known designs, and the higher levels (4–6) involve inverse reasoning.
Level 1 is focused on testing universal knowledge and does not entail queries about any specific design. Level 2 is limited to questions that can be answered from the static description of the design. The higher levels involve reasoning over static and dynamic behaviors of the design. The scope of the problem is closed-world for Levels 1-4 wherein the space of designs and the requirements are given. At Level 5, the synthesis problems have a given set of requirements, the design space is open-world, but subject to reuse or bounded modification constraints. Level 6 is the most competent cognitive level of \name\ which considers an open scope, where design objectives may be underspecified, constraints may conflict, and new performance metrics must be inferred or prioritized.





\label{sec:method}

\section{Evaluation Coverage, Completeness, and Sufficiency}
\label{sec:eval}
The engineering cognition taxonomy discussed above provides an effective framing for organizing evaluation queries, however, a robust evaluation framework for \name\ must demonstrate comprehensive \textit{coverage} across the full breadth of engineering domains, \textit{completeness} in assessing the depth of cognitive capabilities, and \textit{sufficiency} in ensuring that each evaluation scenario tests meaningful and domain-relevant competencies.

To achieve this, we complement the cognitive taxonomy discussed above with a secondary taxonomy that includes aspects of engineering domain coverage. 
This {secondary layer} includes structured metadata tagging along physics-based and engineering-specific dimensions. These tags provide essential contextual and domain-specific information that augments the primary cognitive taxonomy and enables fine-grained filtering, targeted benchmarking, and adaptive evaluation generation. Each tag class supports modularity, domain transfer, and traceability across problem types. Key dimensions include:

\begin{itemize}[leftmargin=*]

    \item \textbf{System Type}: Specifies the application or system context within which the engineering problem is situated. This ensures real-world relevance and contextual grounding. Common types include:
    \begin{itemize}[leftmargin=1em]
        \item \textit{eVTOL systems}: Electric propulsion, rotorcraft dynamics, flight control
        \item \textit{HVAC systems}: Split vs. packaged systems, refrigerant flow, compressor staging
        \item \textit{Aerospace/Spacecraft}: Radiators, attitude control, orbital mechanics
        \item \textit{Energy Systems}: Microgrids, battery packs, solar integration
        \item \textit{Robotics/Mechatronics}: Actuator sizing, kinematic control, onboard computing
    \end{itemize}
    Tagging by system type enables evaluation generation tailored to specific industries or product verticals.

    \item \textbf{Design Scope}: Captures the granularity or scope of the design or analysis task, enabling assessments at different abstraction levels:
    \begin{itemize}[leftmargin=1em]
        \item \textit{Component-level}: Selecting or modeling individual parts like motors, heat exchangers, or valves
        \item \textit{Subsystem-level}: Designing integrated modules like propulsion units, HVAC loops, or powertrains
        \item \textit{System-level}: End-to-end design tasks involving architecture selection, performance trade-offs, and inter-domain reasoning
    \end{itemize}
    This stratification supports progressive evaluation complexity and modular benchmarking.

    \item \textbf{Domain}: Refers to the underlying physical or engineering science involved in the question. Examples include:
    \begin{itemize}[leftmargin=1em]
        \item \textit{Thermal}: Heat transfer, cooling loops, thermodynamic cycles
        \item \textit{Electrical}: Circuits, motors, battery models, signal integrity
        \item \textit{Control}: Feedback loops, PID tuning, stability margins
        \item \textit{Structural}: Stress-strain analysis, beam loading, factor of safety
        \item \textit{Fluid/Airflow}: Propeller flow dynamics, duct pressure drops, fan selection
    \end{itemize}
    These tags allow evaluation sets to be curated by domain, ensuring a diverse and representative distribution across physical modeling spaces.
    
    \item \textbf{Modeling Requirements}: Indicates the analytical or simulation models needed to solve or validate the problem. This helps calibrate \name\ agents capabilities to physics reasoning depth:
    \begin{itemize}[leftmargin=1em]
        \item \textit{Steady-state models}: Static conditions, fixed boundary values (e.g., airflow in a duct at a fixed RPM)
        \item \textit{Transient models}: Time-varying behaviors (e.g., startup thermal lag, motor ramp-up)
        \item \textit{Linear vs. Nonlinear models}: Governing equation complexity (e.g., Ohm’s Law vs. Navier-Stokes)
        \item \textit{Multiphysics models}: Coupled domains (e.g., electro-thermal effects in battery systems)
    \end{itemize}
    This enables evaluations to be generated based on solver or simulator capabilities of a given \name\ system.

    \item \textbf{Applicable Standards}: Refers to engineering norms, certification guidelines, or industry regulations that may constrain or guide the design. This ensures practical realism and regulatory awareness. Common examples include:
    \begin{itemize}[leftmargin=1em]
        \item \textit{AHRI (HVAC)}: Component certification and energy efficiency rating
        \item \textit{ASME (Mechanical Systems)}: Pressure vessel design, code compliance
        \item \textit{UL (Electrical)}: Fire safety, insulation, current limits
        \item \textit{MIL-STD (Defense Systems)}: Shock resistance, environmental qualification
    \end{itemize}
    Standards tagging allows problem constraints to reflect real-world feasibility and industry-aligned requirements.
\end{itemize}

These metadata tags serve not only as classification tools but also as dynamic \textit{control parameters} for adaptive test generation pipelines. By incorporating them into the generation and filtering logic, the system can synthesize and tailor evaluation questions based on domain-specific constraints, cognitive difficulty, or model fidelity requirements. This capability is essential for constructing benchmarks that reflect real-world challenges faced by engineering professionals and allow \name\ systems to be assessed in a principled, scalable manner.
For example, the metadata tags enable:
\begin{itemize}[leftmargin=*]
    \item \textbf{Targeted Sampling}: Queries such as \textit{``generate 10 transient thermal problems for HVAC subsystems''} or \textit{``produce control-system questions focused on drones with nonlinear dynamics''} can be executed by filtering across domain, system type, modeling requirement, and cognitive level.
    \item \textbf{Stratified Evaluation}: Evaluation sets can be balanced or stress-tested by enforcing controlled distributions across tags. For instance, a diagnostic benchmark may intentionally mix linear and nonlinear cases, or alternate between subsystem and full-system tasks to expose weaknesses in generalization.
    \item \textbf{Curriculum-Aligned Progression}: Evaluation sequences can be scaffolded to align with an \name’s developmental milestones or a human educational curriculum. A curriculum designer may start with steady-state component-level problems in the electrical domain (e.g., \textit{``compute resistance of a given circuit''}) and progressively introduce transient, multi-domain, and system-level design tasks (e.g., \textit{``optimize the energy consumption of a hybrid HVAC loop with thermal and control interactions''}).
\end{itemize}

This fine-grained controllability transforms the metadata layer into a powerful lever for shaping the evaluation landscape - not just retroactively tagging problems, but actively guiding their generation to match learning objectives, domain focus, or benchmarking goals.
This dual taxonomy -- combining cognitive levels and domain-specific metadata -- enables \textit{exhaustive yet modular} question generation across a wide variety of engineering domains. Evaluations can be flexibly assembled, diversified, or scaled to meet specific objectives. For example, to benchmark a subsystem-level control strategy for HVAC under transient thermal loads, the evaluation will focus on  relevant Apply- or Analyze-level questions filtered by tags such as \texttt{[HVAC, Subsystem, Thermal, Transient, Control]}.

Each evaluation prompt is synthesized by selecting from structured templates that integrate:
\begin{itemize}[leftmargin=*]
    \item A \textbf{System-level Context} (e.g., ``a drone propulsion module'', ``a refrigeration loop'', or ``a spacecraft radiator''),
    \item A \textbf{Cognitive Target} (e.g., recall, simulate, diagnose, create, reflect),
    \item A \textbf{Task Type} (e.g., swap a component, identify failure mode, optimize for cost, design a system).
\end{itemize}

These elements are combined using reusable semantic structures to yield naturalistic and goal-oriented evaluation prompts that can be generated using LLMs. Examples of such template structures include:
\begin{itemize}[leftmargin=*]
    \item ``Given ---, what happens if ---''
    \item ``Design a system that meets --- constraints''
    \item ``Explain why --- fails under --- conditions''
    \item ``Compare design A and B for performance under ---''
\end{itemize}

Instantiating these templates with different metadata configurations leads to diverse cognitive and domain-specific challenges. For example:
\begin{itemize}[leftmargin=*]
    \item Level 3 (\textit{Apply}): \textit{``Given a 400 Kv motor and 22.2V input, compute the RPM.''}
    \item Level 5 (\textit{Create}): \textit{``Design a propulsion system for a 10 kg drone that minimizes current draw while hovering.''}
    \item Level 6 (\textit{Reflect}): \textit{``You designed a variable refrigerant flow (VRF) HVAC system for a 20-story building located in a hot-humid climate. The design prioritizes energy efficiency by minimizing compressor cycling and optimizing part-load performance using manufacturer data and simulation models. After reviewing your design, a domain expert points out that you may have under-accounted for maintenance constraints and real-world part-load degradation. What modeling or data assumptions might have led to this oversight?''
    }

\end{itemize}

Together, the integration of the primary (cognitive) taxonomy and the secondary (domain and metadata) tagging framework ensures that evaluation quality is upheld along three critical dimensions:

\begin{itemize}[leftmargin=*]
    \item \textbf{Coverage} spans the full breadth of engineering domains, system types, physics phenomena, and modeling regimes. This guarantees that evaluations are not biased toward a particular field (e.g., electrical vs. thermal) or abstraction level (e.g., component vs. system), and instead reflect the interdisciplinary nature of real-world engineering.
    
    \item \textbf{Completeness} ensures that the full cognitive spectrum, from factual recall and conceptual understanding to synthesis, diagnosis, and meta-reasoning, is systematically assessed. Every cognitive level is addressed across a diverse set of contexts, enabling evaluation of both isolated skills and integrative reasoning.

    \item \textbf{Sufficiency} is achieved by grounding each question in realistic engineering scenarios, supported by domain-specific constraints, physics-based models, and industry standards. Evaluation tasks are not merely academic, but they are structured to reflect challenges that would be encountered by human engineers in practical design, analysis, or troubleshooting situations.
\end{itemize}

By distributing these three qualities -- coverage, completeness, and sufficiency -- across cognitive levels, domain contexts, task types, modeling complexity, and design granularity, this framework supports the systematic construction of rigorous, balanced, and extensible evaluation banks. These evaluations can adapt to the evolving capabilities of \name\ systems and provide meaningful benchmarks for reasoning, generalization, and innovation across engineering disciplines.

\section{Application to Propeller-Motor Matching Problem}
To ground our approach to evaluating \name, we chose an electric vertical takeoff and landing (eVTOL) aircraft as a representative example. This choice is deliberate: the problem's multi-physics nature and the complex dependencies of design decisions create a realistic and nuanced scenario well-suited for \name\ evaluation. Additionally, the general familiarity and broad accessibility of drone domain knowledge mean that even readers without specific expertise can readily grasp the evaluation queries, and facilitate a comparative analysis of performance across human engineers, general purpose LLMs, and customized LLMs.  

Specifically, for this discussion we limit our scope to a propellor-motor sizing and matching problem for an evTOL aircraft.  This design task involves selecting an appropriate combination of electric motor and propeller such that the overall propulsion system can generate sufficient thrust to lift and maneuver the vehicle, while maintaining efficiency, weight, thermal compliance, and structural robustness.
The design challenge spans multiple physical domains:
\begin{itemize}[leftmargin=*]
\item Electrical domain: Converting battery power to motor torque (motor voltage, current, Kv rating).
\item Mechanical domain: Transmitting torque as angular rotation to the propeller.
\item Aerodynamic domain:  Thrust as a function of propeller pitch, airfoil geometry, diameter, and RPM.
\end{itemize}

The typical requirements might include hover and cruise thrust levels, battery and motor constraints (e.g., voltage, current), thermal limits, physical design space, and redundancy constraints.
Addressing these requirements involves understanding of a tractable but realistic multiphysics system, requires both static and dynamic behavior reasoning capabilities, and exploration over both open and closed-world design spaces. We present examples of question-answer aligned with each \name\ level with explanation about the corresponding cognitive level in this section. 
Levels 2-6 require a design and/or requirements with respect to which understanding, applying, analyzing designs are performed. 

It's important to understand that while we've used textual responses to expected questions in this document for ease of reading, real-world applications would generate structured artifacts and design representations compatible with engineering tools like CAD, SysML, and Modelica. Evaluating these outputs would then necessitate syntactic and semantic checks using engineering software. Accordingly, the evaluation framework need to be be highly pluggable and customizable, but a detailed discussion of these aspects is beyond the scope of this current explanation.

We use the following specific example design problem for questions and answers. We present one example for each cognition level below and provide more examples in Appendix~\ref{app:evtol}.  

\noindent \textbf{Example Design}
A quadrotor eVTOL drone designed for urban logistics, with a 
max takeoff weight: 12 kg (including 5 kg payload), 
power source: 6S (22.2V) LiPo battery, 
motors: 380 Kv, limited to 25A max draw per motor, 
propellers: 18x6 fixed-pitch, and
configuration: 4 motors. 
The design is tested under standard sea-level conditions (air density = 1.225 kg/m³).\\
\\
\textbf{Metatag:} eVTOL System, Multiphysics Model, Electrical/Control/Aerodynamics Domain.\\


\begin{flushleft}
\hrule\medskip

\textbf{Level 1 – Remember}\par
\textbf{Question:} What is the thrust equation for a propeller in static air?\\
\textbf{Expected answer:} 
$T = C_T \cdot \rho \cdot n^2 \cdot D^4$ \\
\textbf{Justification:} Tests recall of fundamental physics and domain-specific equations.

\medskip\hrule\medskip

\textbf{Level 2 – Understand}\par
\textbf{Question:} What are the propeller parameters in this design?\\
\textbf{Expected answer:}
The propeller diameter is 18 inches, the pitch is 6 inches, and the type is fixed pitch. \\
\textbf{Justification:} Tests understanding of propeller parameters in the design context.

\medskip\hrule\medskip

\textbf{Level 3 – Apply}\par
\textbf{Question:} In the example design, compute the thrust if we increase the propeller diameter to 20 in while keeping RPM constant.\\
\textbf{Expected answer:} 
If RPM remains constant, increasing propeller diameter from $18''$ to $20''$ results in approximately $52\%$ more thrust per motor. \\
\textbf{Justification:} Tests conceptual grasp of the thrust–diameter relationship.

\medskip\hrule\medskip

\textbf{Level 4 – Analyze}\par
\textbf{Question:} Why could the drone’s thrust be insufficient at a 12 kg take-off weight if the loaded RPM is 7,500?\\
\textbf{Expected answer:} At 7,500 RPM, the 18x6 propellers will not generate sufficient thrust to lift the 12 kg eVTOL. The required thrust per motor is 29.4N and the thrust available at 7500 RPM is 26.4N. \\
\textbf{Justification:} Tests sensitivity-analysis reasoning for aerodynamic performance.

\medskip\hrule\medskip

\textbf{Level 5 – Create}\par
\textbf{Question:} Design a quadrotor eVTOL that can carry a 6 kg payload (MTOW = 14 kg), hover $\geq$ 12 min, and keep each motor $\leq$ 22 A at 22.2 V. Recommend motor Kv and propeller size.\\
\textbf{Expected answer:} 
Recommended Setup. 
Motor: 340 Kv (high-torque, low-RPM), 
Propeller: 20×6 inch fixed-pitch, 
Voltage: 6S (22.2V), 
Thrust per motor: 34.3N, 
Estimated Current Draw: 19–21A per motor, 
Expected Hover Time (using 12,000 mAh battery): 12–14 minutes. This setup balances thrust, efficiency, and current limits. Large props 
reduce disk loading and improve endurance. \\
\textbf{Justification:} Synthesizes a configuration satisfying performance and electrical constraints.

\medskip\hrule\medskip

\textbf{Level 6 – Reflect}\par
\textbf{Question:} 
You designed a propulsion system for a 5 kg fixed-wing drone. The motor-propeller pair you selected was optimized for cruise efficiency using standard performance maps from the manufacturer. During flight testing, the drone exhibited inadequate climb performance and marginal takeoff thrust under high-altitude conditions. What assumptions in your design or modeling may have contributed to the underperformance? \\
\textbf{Expected answer:} 
I relied on manufacturer-provided propeller and motor performance maps, which are typically derived under sea-level standard temperature and pressure. These maps do not account for the significant drop in air density at higher altitudes, which directly reduces propeller thrust generation. The simulation pipeline did not include environmental variation or altitude-corrected performance prediction. I failed to simulate or verify performance across the full flight envelope, particularly under high-altitude, low-density scenarios. 
I could also have scaled thrust performance with altitude-adjusted air density. \\
\textbf{Justification:} 
This response shows
causal reasoning (understanding why the issue occurred), 
introspection (identifying what data or modeling assumptions were missing), 
design methodology critique (reflecting on the structure of the decision-making process), and
proactive correction (suggesting specific workflow and modeling improvements).

\medskip\hrule
\end{flushleft}

\label{sec:propMotor}

\section{Automated Evaluation and Alignment with Human Expertise}
\label{sec:auto}

A central challenge in design of an evaluation framework for evaluating \name\ agents is balancing \textit{automation of scoring} with the need for \textit{human-level understanding} in higher-order tasks. The framework needs to enable scalable, structured evaluation aligned with expert judgment, while automating as much of the evaluation process as possible. We consider these challenges and approaches in context of the cognition taxonomy discussed earlier. 

\textbf{Levels 1 to 3} (\textit{Remember, Understand, Apply}) are largely objectively scorable. Responses can be evaluated through symbolic math solvers, reasoning engines, lookup tables,  knowledge bases, and domain-specific simulation models (e.g., computing thrust or efficiency).
These levels typically involve deterministic or well-constrained answers, making full automation feasible.

\textbf{Level 4} (\textit{Analyze}) introduces diagnostic reasoning. Here, responses may include fault localization (e.g., ``identify the broken component'') and design fixes (e.g., ``suggest a redesign to resolve the issue''). While localization may yield multiple correct answers, the proposed fix can be validated by patching the system and simulating its behavior to confirm resolution.

\textbf{Level 5} (\textit{Create}) involves open-ended design synthesis. Automated scoring here focuses on partial correctness and feasibility, such as
verifying that design constraints are satisfied (e.g., power, weight, voltage), ensuring physical consistency across components (e.g., torque-speed compatibility), and comparing solutions against reference architectures or simulation baselines.
Simulation-based validation or optimization metrics (e.g., Pareto dominance) are used to assess creativity and effectiveness.

\textbf{Level 6} (\textit{Reflect}) 
At Level 6, an \name\ demonstrates meta-cognitive capabilities by reflecting on its own reasoning processes, knowledge boundaries, modeling assumptions, and the limitations of its outputs. Rather than just evaluating the quality of designs or making trade-offs, the agent critically examines the soundness, reliability, and completeness of its own decisions and design workflows.
This level focuses on the agent's capability to 
identify gaps or flaws in its prior reasoning or simulations, 
recognize when additional data, expertise, or validation is needed,
evaluate whether it has used appropriate methods and whether results are valid under non-ideal or out-of-distribution conditions, and 
articulate the limitations of its knowledge, tools, or models.
These often require human-in-the-loop scoring to assess the open-world reasoning. 
However, partial automation is viable through heuristics such as 
scoring argument consistency with design goals, and augmenting LLMs used as a judge with additional information such as  the expected knowledge or reasoning gaps in the eAGI. 


By aligning scoring methods with the cognitive complexity of evaluation tasks, our proposed framework establishes a principled and scalable mechanism for assessing \name\ performance. Each tier of cognitive difficulty is matched with an appropriate evaluation strategy, ensuring that the assessment process remains both rigorous and computationally tractable. Specifically, the evaluation uses the following:

\begin{itemize}[leftmargin=*]
    \item \textbf{Objective scoring} is prioritized for lower cognitive levels (e.g., \textit{Remember}, \textit{Understand}, and \textit{Apply}). These tasks often involve deterministic answers, such as recalling equations, computing values, or applying rules, which can be graded automatically using symbolic evaluation, lookup tables, or numeric simulation. This enables fast, repeatable assessments at scale without requiring human involvement.
    
    \item \textbf{Simulation-augmented heuristics} are employed at intermediate levels (e.g., \textit{Analyze} and \textit{Create}), where responses include failure localization, subsystem redesign, or design synthesis. While these answers may not be uniquely defined, their soundness can be verified by simulating the impact of the proposed fix or design. Partial credit can be awarded based on constraint satisfaction, consistency across subsystems, or proximity to optimal reference solutions.
    
    \item \textbf{Expert-in-the-loop review and agent as a judge} is reserved for high-level cognitive tasks (e.g., \textit{Reflect}) that require nuanced judgment, such as evaluating trade-offs, critiquing alternative designs, or reasoning about uncertainty and system limitations. These tasks often depend on implicit domain knowledge or creativity, where correctness is not easily reducible to a numeric score. In such cases, human evaluators provide qualitative assessment or validation based on domain expertise, and LLM agents have been recently demonstrated to effectively surrogate human evaluators in some settings \cite{zhuge2024agent}.
\end{itemize}

This tiered scoring framework enables automated, scalable, and repeatable benchmarking of \name\ capabilities.


\section{Conclusion}
In this paper, we have discussed a novel evaluation framework for \name\, grounded in Bloom’s taxonomy and tailored to the complexities of engineering design. We introduced a six-level cognitive hierarchy specifically adapted for engineering tasks and demonstrated how it enables structured, scalable, and cognitively aligned question generation.

Our proposed approach combines reusable question templates, domain-specific design contexts, and physics-aware metadata tagging to generate diverse and rigorous evaluation queries. By anchoring assessments in real-world engineering reasoning from basic recall to advanced design synthesis and reflective critique, our framework enables principled benchmarking of \name\ agents across a wide range of physical systems and engineering domains.

This taxonomy-driven methodology not only ensures broad coverage and deep cognitive progression but also supports automated and semi-automated scoring through symbolic evaluation, simulation-based validation, expert-in-the-loop review, LLM-as-a-judge~\cite{zheng2023judging}, and Agent-as-a-judge~\cite{zhuge2024agent}. Ultimately, this work provides a foundation for the systematic development, benchmarking, and comparison of AI systems capable of collaborating in complex engineering workflows.

While our framework offers a structured path toward scalable and cognitively grounded evaluation, several limitations remain, particularly at the higher end of the cognitive spectrum. These challenges point to key opportunities for future research:

\begin{itemize}[leftmargin=*]
    \item {Automation bottlenecks at advanced cognitive levels}: Tasks involving open-ended design, innovation, or trade-off reasoning (e.g., Levels 5 and 6) still require expert judgment. Although simulation and constraint-checking can partially validate responses, fully automating the scoring of creativity, novelty, or rationale remains an unsolved challenge. LLM-as-a-judge and agent-as-a-judge partially address this challenge, but their effectiveness in the evaluation of \name\ has not been demonstrated yet. 

    \item {Absence of common evaluative frameworks for subjective reasoning}: Reflective responses (e.g., ``Which design is better and why?'') lack shared semantic representations, making it difficult to ensure that automated scoring aligns with expert intuition or domain-specific reasoning practices. Learning reward models to capture human preferences can partially address this challenge. 

    \item {Limited traceability of multi-step reasoning processes}: \name\ responses to complex design tasks often involve multiple implicit reasoning steps. Without interpretable intermediate traces, it is difficult to assess whether solutions stem from principled reasoning or surface-level heuristics.

    \item {Sensitivity to domain-specific constraints and complex dynamics}: Certain domains, such as structural reliability or control stability, exhibit highly nonlinear behaviors that complicate automated evaluation. Robust scoring in these contexts will require tighter coupling with high-fidelity simulation environments and formal verification techniques.
\end{itemize}

Improving automation, explainability, and fidelity in these areas will be essential for enabling robust, continuously improving benchmarks. As \name\ approach expert-level reasoning, the evaluation infrastructure must evolve to not only track their progress but also challenge them with the same rigor expected in professional engineering practice.

\label{sec:conc}

\bibliographystyle{plainnat}
\bibliography{engbench}

\newpage
\appendix

\section{Example Evaluation Questions for Propeller-Motor Matching}
\label{app:evtol}
\noindent \textbf{Example Design}
A quadrotor eVTOL drone designed for urban logistics, with a 
max takeoff weight: 12 kg (including 5 kg payload), 
power source: 6S (22.2V) LiPo battery, 
motors: 380 Kv, limited to 25A max draw per motor, 
propellers: 18x6 fixed-pitch, and
configuration: 4 motors. 
The design is tested under standard sea-level conditions (air density = 1.225 kg/m³).\\
\\
\textbf{Metatag:} eVTOL System, Multiphysics Model, Electrical/Control/Aerodynamics Domain.

\textbf{Level 1:  Remember}  \\
\\
\textbf{Question:} What is the thrust equation for a propeller in static air? \\
\textbf{Expected Answer:}  
$T = C_T \cdot \rho \cdot n^2 \cdot D^4$ \\
 \textbf{Justification: }
Tests recall of fundamental physics and domain-specific equations. \\
\\
\textbf{Question:} What does motor Kv represent? \\
\textbf{Expected Answer:}  
RPM per Volt under no-load conditions \\ 
\textbf{Justification:}  
Tests knowledge of motor specs used in design decisions. \\
\\
\textbf{Question:} What is the relationship between motor Kv and RPM? \\
\textbf{Expected Answer:}  
RPM = Kv × Voltage (under no-load conditions) \\
\textbf{Justification:}  
Tests factual recall of domain-specific equations. \\
\\
\textbf{Question:} What voltage corresponds to a 6S LiPo battery? \\
\textbf{Expected Answer:}  
Approximately 22.2V (6x3.7V) \\
 \textbf{Justification: }
Tests standard component knowledge from the domain. \\
\\
\textbf{Level 2:  Understand} \\
\\
\textbf{Question:} What are the propeller parameters in this design? \\
\textbf{Expected Answer:}  
The propeller diameter is 18 inches, the pitch is 6 inches, and the type is fixed pitch. \\
 \textbf{Justification: }
Tests understanding of propeller parameters in the context of a design. \\
\\
\textbf{Question:} What is the voltage range and the nominal voltage for the design?\\
\textbf{Expected Answer:}  
The voltage range is 19.8V - 25.2V and the nominal voltage is  22.2V \\
 \textbf{Justification: }
Tests understanding of battery and motor voltage parameters in the design. \\
\\
\textbf{Question:} What are the key parameters in the design that determine the maximum torque?
\\
\textbf{Expected Answer:}  
The maximum torque in this design is primarily governed by the motor Kv, current limit (25A), and the torque constant. \\
 \textbf{Justification: }
Tests understanding of the torque equation in the context of this design where the ESC capability and thermal limits are not specified.  \\
\\
\textbf{Level 3: Apply}  \\
\\
\textbf{Question:} In the example design, compute the thrust if we increase the propeller diameter to 20 inches while keeping RPM constant? \\
\textbf{Expected Answer:}  
If RPM remains constant, increasing propeller diameter from $18''$ to $20''$ results in approximately $52\%$ more thrust per motor. \\
 \textbf{Justification: }
Tests conceptual grasp of thrust-diameter relationship in the design context. \\
\\
\textbf{Question:} What is the expected RPM of the 380 Kv motor at 22.2V used in this eVTOL? \\
\textbf{Expected Answer:}  
380x22.2 = 8436 RPM \\
 \textbf{Justification: }
Applies a direct formula with design-specific parameters. \\
\\
\textbf{Question:} How would increasing the motor Kv from 380 to 420 in this drone affect the torque output? \\
 \textbf{Expected Answer:}  
Increasing Kv from $380$ to $420$ reduces the torque constant 
Kt from approximately $0.0251$ Nm/A to $0.0227$ Nm/A.
At $25A$, torque drops from $0.63$ Nm ($380$ Kv) to $0.57$ Nm ($420$ Kv), a $10\%$ reduction in torque. \\
 \textbf{Justification: }
Tests understanding of the relationship between Kv, Kt and torque in the design. \\
\\
\textbf{Level 4: Analyze} \\
\\
\textbf{Question:} Why could the drone's thrust be insufficient at 12 kg takeoff weight if the loaded RPM is 7500? \\
\textbf{Expected Answer:}  
At 7,500 RPM, the 18x6 propellers will not generate sufficient thrust to lift the 12 kg eVTOL. The required thrust per motor is 29.4N and the thrust available at 7500 RPM is 26.4N. \\
 \textbf{Justification: }
Tests sensitivity analysis understanding for aerodynamic performance. \\
\\
\textbf{Question:} To increase thrust by 10\% in this system without changing motor or battery, what propeller modification would you suggest? \\
\textbf{Expected Answer:}  
Increase pitch to 18x7 or diameter to 19 inches, if structurally feasible. \\
 \textbf{Justification: }
Applies performance tuning logic while maintaining constraints.
This eVTOL draws $>$15A per motor during hover. \\
\\
\textbf{Question:} What might be the cause and how can this be reduced? \\
\textbf{Expected Answer:}  
Switching to a lower Kv motor (e.g., 340–360 Kv) or 
using larger, lower-pitch propellers can reduce current draw. \\
 \textbf{Justification: }
Diagnoses the current overdraw from configuration parameters. \\
\\
\textbf{Question:} The 18x6 props struggle during vertical climb. Suggest design changes. \\
\textbf{Expected Answer:}  
Increase pitch or RPM, reduce vehicle weight, or change to 20x6 props. \\
 \textbf{Justification: }
Requires targeted reasoning to improve climb performance. \\
\\
\textbf{Question:} The drone vibrates excessively at high RPM. What design issue might be causing this? \\
\textbf{Expected Answer:}  
Possible prop imbalance or structural resonance with harmonics. \\
 \textbf{Justification: }
Applies failure diagnosis to aerodynamic/mechanical systems. \\
\\
\textbf{Level 5: Create}  \\
\\
\textbf{Question:} Design a quadrotor eVTOL capable of carrying a 6 kg payload with a maximum takeoff weight of 14 kg. The drone must hover for at least 12 minutes, and each motor should not exceed 22A current draw at nominal voltage (22.2V). Recommend suitable motor Kv and propeller size. \\
\textbf{Expected Answer:}  
Recommended Setup. 
Motor: 340 Kv (high-torque, low-RPM), 
Propeller: 20×6 inch fixed-pitch, 
Voltage: 6S (22.2V), 
Thrust per motor: 34.3N, 
Estimated Current Draw: 19–21A per motor, 
Expected Hover Time (using 12,000 mAh battery): 12–14 minutes. This setup balances thrust, efficiency, and current limits. Large props 
reduce disk loading and improve endurance. \\
 \textbf{Justification: }
Synthesizes a new configuration from performance and electrical constraints. \\
\\
\textbf{Question:} Design an eVTOL for medical payload delivery in hilly terrain. Requirements include:
Payload: 4 kg, 
Max Takeoff Weight: 11 kg, 
Silent operation for urban environments, 
Hover time: 10 minutes minimum, 
Must fit within a compact 0.7m x 0.7m footprint \\
\textbf{Expected Answer:}  
Recommended Setup. 
Motor: 320 Kv, 
Propeller: 16×5.4 inch carbon fiber low-noise props, 
Voltage: 6S (22.2V), 
Prop Layout: Coaxial quadrotor (8 props in total to fit within tight space while maintaining thrust), 
Performance Estimate:
Required per-prop thrust: 13.5 N, 
Operating RPM: 7000–7500, 
Current draw: ~17–20A per motor. 
Coaxial configuration enables high thrust in a constrained footprint. Low-Kv motors with quiet props reduce noise signature while meeting endurance and lift targets. \\
 \textbf{Justification: }
Synthesizes a new configuration from the given performance and geometry constraints and also satisfy preferences such as silent operation.\\
\\
\textbf{Question:} Design a research eVTOL drone to operate in high-altitude, thin-air conditions (up to 3,000 meters). It should carry a 3 kg sensor payload, total weight: $\leq 9$ kg, 
maintain stable hover in air density 0.9 $kg/m^3$, and
use standard 6S battery (22.2V). Produce this design in the SysML format.\\
\textbf{Expected Answer:}  

\begin{lstlisting}[mathescape=true]
@startuml DroneRequirements
' SysML Requirements Diagram for High-Altitude Research eVTOL Drone
left to right direction
package "High-Altitude eVTOL Drone Requirements" {
  requirement R1 <<requirement>> {
    Id = R1
    Text = "Shall carry 3 kg sensor payload"
  }
  requirement R2 <<requirement>> {
    Id = R2
    Text = "Maximum take-off weight (MTOW) $\leq$ 9 kg"
  }
  requirement R3 <<requirement>> {
    Id = R3
    Text = "Shall maintain stable hover at air density $\rho$ = 0.9 kg/m^3 (3000 m ASL)"
  }
  requirement R4 <<requirement>> {
    Id = R4
    Text = "Shall use standard 6 S Li-ion/Li-Po battery (nominal 22.2 V)"
  }
  requirement R5 <<requirement>> {
    Id = R5
    Text = "Hover endurance $\geq$ 10 min at MTOW"
  }
  requirement R6 <<requirement>> {
    Id = R6
    Text = "Operational ceiling $\geq$ 3000 m"
  }
}
@enduml

@startuml DroneBDD
' SysML Block Definition Diagram
skinparam rectangle {
  BackgroundColor Lavender
}
block "Research eVTOL Drone" as Drone {
  +mass : kg = 9
  +endurance_hover : min = 12
  +operational_ceiling : m = 3000
}

block Airframe {
  +material = CFRP
  +arm_length : mm = 450
  +frame_mass : kg = 0.9
}

block PropulsionSubsystem {
  +motor_model = "400 kV BLDC"
  +prop_diameter : inch = 18
  +esc_rating : A = 60
  +thrust_per_rotor : N = 22.5
  +rotor_mass : kg = 0.5
}

block PowerSubsystem {
  +battery_type = "6 S Li-ion"
  +capacity : Ah = 10
  +voltage_nom : V = 22.2
  +max_current : A = 80
  +battery_mass : kg = 2
}

block Avionics {
  +flight_controller = "Pixhawk 6X"
  +GNSS = "RTK-GNSS"
  +avionics_mass : kg = 0.35
}

block SensorPayload {
  +payload_mass : kg = 3
  +interfaces = "USB-C, ETH"
}

Drone *-- "4" PropulsionSubsystem : composed_of
Drone *-- PowerSubsystem
Drone *-- Airframe
Drone *-- Avionics
Drone *-- SensorPayload
@enduml

@startuml DroneIBD
' SysML Internal Block Diagram
block "Research eVTOL Drone" {
  part airframe : Airframe
  part pwr : PowerSubsystem
  part flightCtl : Avionics
  part sensor : SensorPayload
  part prop[4] : PropulsionSubsystem

  pwr.power_out -> flightCtl.power_in : 22.2 V DC
  pwr.power_out -> prop.power_in
  flightCtl.cmd_out -> prop.cmd_in : PWM
  flightCtl.data_bus <--> sensor.data_bus : CAN/ETH
}
@enduml

@startuml DroneParam
' SysML Parametric Diagram - Hover \& Endurance Performance
block Drone
constraint "HoverThrust" as c1 {
  {T = m * g}
}
constraint "HoverPower" as c2 {
  {P = (T^(3/2)) / (sqrt(2*\rho*A_total) * \eta)}
}
constraint "Endurance" as c3 {
  {t = (C * V * $\eta_{batt}$) / P}
}
value m : kg = 9
value g : m/s^2 = 9.81
value $\rho$ : kg/m^3 = 0.9
value A_total : m^2 = 0.636
value $\eta$ : dimensionless = 0.7
value C : Ah = 10
value V : V = 22.2
value $\eta_{batt}$ : dimensionless = 0.95
derived T : N
derived P : W
derived t : min

m -> c1
g -> c1
c1 -> c2.T
$\rho$ -> c2
A_total -> c2
$\eta$ -> c2
c2 -> c3.P
C -> c3
V -> c3
$\eta_{batt}$ -> c3
@enduml
\end{lstlisting}
 \textbf{Justification: }
Synthesizes a new design from the given performance and electrical requirements to satisfy the constraints on the operation requirement. \\
\\
\textbf{Level 6: Reflect} \\
\\
\textbf{Question:} 
You designed a propulsion system for a 5 kg fixed-wing drone. The motor-propeller pair you selected was optimized for cruise efficiency using standard performance maps from the manufacturer. During flight testing, the drone exhibited inadequate climb performance and marginal takeoff thrust under high-altitude conditions. What assumptions in your design or modeling may have contributed to the underperformance? \\
\textbf{Expected answer:} 
I relied on manufacturer-provided propeller and motor performance maps, which are typically derived under sea-level standard temperature and pressure. These maps do not account for the significant drop in air density at higher altitudes, which directly reduces propeller thrust generation. The simulation pipeline did not include environmental variation or altitude-corrected performance prediction. I failed to simulate or verify performance across the full flight envelope, particularly under high-altitude, low-density scenarios. 
I could also have scaled thrust performance with altitude-adjusted air density. \\
\textbf{Justification:} 
This response shows
causal reasoning (understanding why the issue occurred), 
introspection (identifying what data or modeling assumptions were missing), 
design methodology critique (reflecting on the structure of the decision-making process), and
proactive correction (suggesting specific workflow and modeling improvements). \\
\\
\textbf{Question:} You recently completed a propeller-motor match for a quadrotor using lithium polymer batteries. Now you are tasked with designing a similar system for a fixed-wing drone using lithium-sulfur batteries. What aspects of your previous design process or assumptions may not transfer?
\\
\textbf{Expected answer:}  
Several aspects of my previous design process for the quadrotor propulsion system may not transfer directly to the fixed-wing drone using lithium-sulfur (Li–S) batteries, due to both platform and energy source differences. Below, I reflect on the key assumptions and design heuristics that require re-evaluation:
\begin{enumerate}[leftmargin=*]
\item Platform Dynamics and Mission Profile:
Quadrotors and fixed-wing drones have fundamentally different propulsion needs. The quadrotor design emphasized hover thrust, symmetric loading across multiple rotors, and rapid throttle response. In contrast, a fixed-wing drone relies primarily on sustained forward thrust with relatively stable loading, optimized for cruise and glide performance. Assumptions about required thrust margins, dynamic load behavior, and motor responsiveness must be revised accordingly.

\item Battery Discharge Characteristics and Energy Density:
My previous design assumed lithium polymer (LiPo) batteries with high discharge rates (20C–60C) and relatively flat voltage curves. Li–S batteries, while offering higher specific energy, have lower discharge rates and a steeper voltage drop-off under load. This affects motor selection (peak current draw must be capped to avoid brownouts or cell damage), 
propeller sizing (efficiency must be prioritized to reduce sustained current draw), and 
power margin assumptions (excess thrust design margins from the quadrotor case are not viable here without compromising endurance).

\item Mass and Volume Constraints:
Li–S batteries, while lighter for the same energy content, have different form factors and may impose new constraints on placement, CG (center of gravity) balancing, and cooling. My prior motor-propeller-battery configuration assumed compact LiPo packs that could be distributed evenly across a symmetrical frame, which does not apply to a fixed-wing fuselage.

\item Cooling and Thermal Assumptions:
In quadrotors, airflow across motors and ESCs (electronic speed controllers) is naturally higher due to rotor wash and hover conditions. Fixed-wing configurations may have less direct airflow, especially during low-speed climb or gliding. The assumption of passive cooling adequacy must be revisited.

\item Performance Maps and Efficiency Models:
My earlier design used performance maps and empirical efficiency curves calibrated under high-discharge LiPo conditions. These may not remain valid under the flatter discharge and dynamic loading behavior of Li–S batteries. I would need to recalibrate simulation models or re-derive efficiency curves specific to the new chemistry.

\item Safety and Degradation Modeling:
Li–S chemistry is less mature than LiPo, with different failure modes, cycle life characteristics, and thermal stability. My original design pipeline lacked battery aging or risk modeling. For long-duration missions, this could lead to overestimating reliability or underestimating maintenance needs.

\end{enumerate}

To ensure transferability, I would
introduce battery-specific electrical and thermal models into the simulation pipeline, 
redefine design objectives to balance peak power constraints with energy utilization over time,
validate propulsion design across multiple mission phases with Li–S discharge curves, and include battery degradation modeling in endurance estimation.
Overall, while prior experience provides useful architectural intuition, many assumptions related to platform dynamics, power availability, and thermal management must be revisited or re-validated in the fixed-wing, Li–S context.\\
\textbf{Justification:} 
This response demonstrates
deep understanding of cross-domain transfer risks, 
ability to reason about interdependencies between components and mission profiles, and 
reflection on implicit assumptions and how to adapt methodology accordingly.
\\
\\
\textbf{Question:} Your propeller-motor design was validated using a low-fidelity thrust model assuming constant efficiency. However, wind tunnel data later shows a significant mismatch in dynamic thrust behavior during acceleration and descent. What simplifications in your simulation contributed to the mismatch?
\\
\textbf{Expected answer:}  \\
The mismatch observed between my predicted thrust performance and the wind tunnel data can be traced back to several simplifications and assumptions in my low-fidelity simulation model. Specifically, the following modeling choices likely contributed to the discrepancy:
\begin{enumerate}[leftmargin=*]
\item Constant Propeller Efficiency Assumption:
I assumed a constant propeller efficiency value across all operating conditions, derived from steady-state cruise data. This simplification ignored how efficiency varies with dynamic factors such as changes in angle of attack, inflow velocity, RPM transients, and air density. In reality, propeller efficiency decreases significantly during unsteady conditions like rapid throttle changes or deceleration phases.

\item Steady-State Thrust Estimation:
My model was based on static or quasi-static thrust coefficients without accounting for unsteady aerodynamics. This means the effects of blade vortex interaction, delayed flow separation, or wake interference during acceleration or descent were not captured. These transient behaviors can reduce instantaneous thrust output relative to predictions made under steady assumptions.

\item Absence of Inflow Modeling:
I did not simulate the variation in inflow velocity at the propeller disc during vertical climb or descent. In these conditions, especially during descent (where propeller blades may enter their own wake—known as the vortex ring state), actual thrust production can drop significantly, which was not reflected in my simplified model.

\item No Coupled Airframe-Propulsion Interaction:
My model did not include aerodynamic feedback from the airframe or its influence on the local flow field around the propeller. In practice, airframe-induced turbulence or flow deflection can reduce effective thrust, particularly in complex maneuvers.

\end{enumerate}

Ultimately, the issue stemmed not from an isolated parameter error, but from a suite of simplifications that collectively failed to capture the system’s behavior under non-steady conditions—highlighting the limits of low-fidelity modeling in dynamic regimes.
\\
\textbf{Justification:} 
This response demonstrates
specific and multi-layered diagnosis of modeling assumptions, and 
deep understanding of aerodynamic and electromechanical coupling. 

\end{document}